\documentclass[letterpaper, 10 pt, conference]{ieeeconf}  

\IEEEoverridecommandlockouts                              
\usepackage{paralist}
\usepackage{xspace}
\usepackage{graphicx}
\usepackage{algorithm}
\usepackage{algpseudocode}
\usepackage{comment}
\usepackage{wrapfig}
\usepackage{float}
\usepackage[table]{xcolor}

\usepackage[dvipsnames]{xcolor}
\usepackage{lipsum}

\usepackage[dvipsnames]{xcolor}

\definecolor{iccvblue}{rgb}{0.21,0.49,0.74}
\usepackage[pagebackref,breaklinks,colorlinks,allcolors=iccvblue]{hyperref}
\usepackage[utf8]{inputenc} 
\usepackage[T1]{fontenc}    
\usepackage{amsmath}        
\usepackage{hyperref}       
\usepackage{url}            
\usepackage{booktabs}       
\usepackage{amsfonts}       
\usepackage{nicefrac}       
\usepackage{microtype}      
\usepackage{caption}        
\usepackage{xcolor}         
\title{Fiducial Exoskeletons: Image-Centric Robot State Estimation}

\newcommand{\projecturl}{%
  \href{https://cameronosmith.github.io/fiducial_exoskeleton/}{
  cameronosmith.github.io/fiducial\_exoskeleton
  }
}

\author{
    Cameron Smith$^{1}$ \quad
    Basile Van Hoorick$^{2}$ \quad
    Vitor Guizilini$^{2*}$ \quad
    Yue Wang$^{1*}$ \vspace{2mm} \\ 
    $^1$USC Physical Superintelligence (PSI) Lab~~~
    $^2$Toyota Research Institute~~~
    \textit{*Equal Advising}
    \\
    \projecturl
}

\begin{document}

\maketitle

\begin{figure*}[t!]
  \centering
  \includegraphics[width=\textwidth]{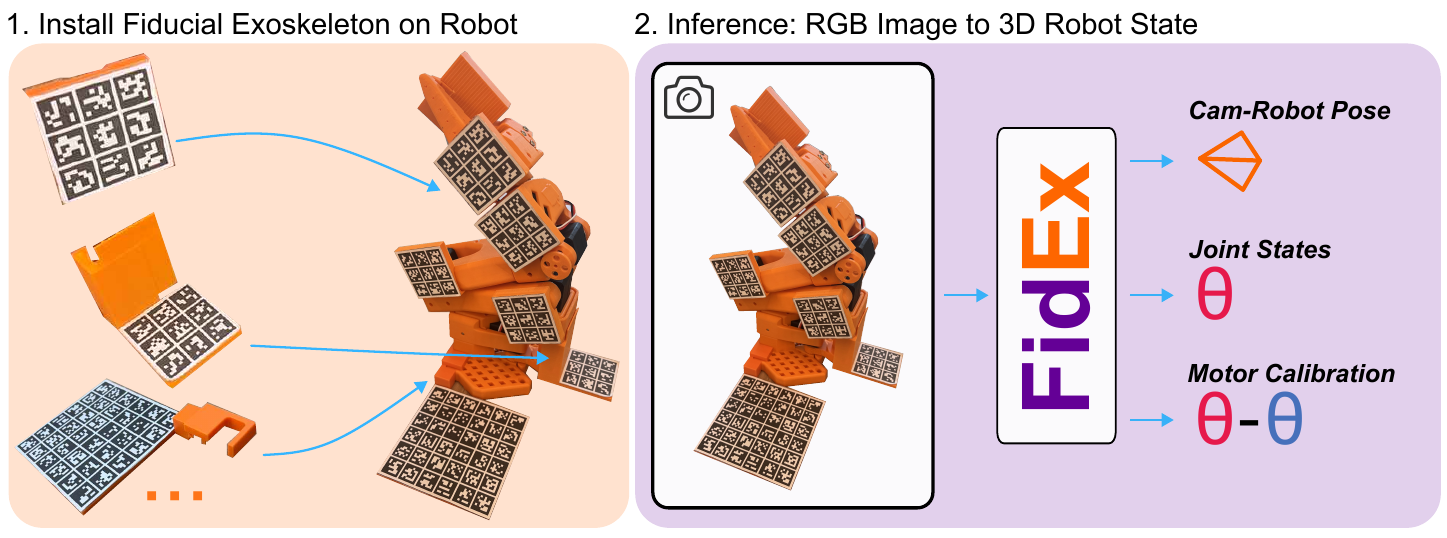}
   \caption{
   \textbf{FidEx Overview}. After installing a lightweight 3D-printed fiducial exoskeleton --- one marker-equipped piece per link ---  FidEx enables instant 3D state estimation from a single RGB image. Given one view, we recover the external camera pose (robot base pose), the full joint state by optimizing over detected 6D link poses, and the joint calibration by aligning optimized joint angles with raw motor readings.
    }
    \label{fig:challenges}
\end{figure*}

\begin{abstract}
We introduce Fiducial Exoskeletons, an image-based reformulation of 3D robot state estimation that replaces cumbersome procedures and motor-centric pipelines  with single-image inference. Traditional approaches—especially robot–camera extrinsic estimation—often rely on high-precision actuators and require time-consuming routines such as hand–eye calibration. In contrast, modern learning-based robot control is increasingly trained and deployed from RGB observations on lower-cost hardware.

Our key insight is twofold. First, we cast robot state estimation as 6D pose estimation of each link from a single RGB image: the robot–camera base transform is obtained directly as the estimated base-link pose, and the joint state is recovered via a lightweight global optimization that enforces kinematic consistency with the observed link poses (optionally warm-started with encoder readings). Second, we make per-link 6D pose estimation robust and simple—even without learning—by introducing the fiducial exoskeleton: a lightweight 3D-printed mount with a fiducial marker on each link and known marker–link geometry.

This design yields robust camera–robot extrinsics, per-link SE(3) poses, and joint-angle state from a single image, enabling robust state estimation even on unplugged robots. Demonstrated on a low-cost (\$100) robot arm, fiducial exoskeletons substantially simplify setup while improving calibration, state accuracy, and downstream 3D control performance. We release code and printable hardware designs to enable further algorithm–hardware co-design.

\end{abstract}

\section{Introduction}
Accurate 3D state estimation and precise 3D control are fundamental components of robotics.
The dominant paradigm for both is to use forward kinematics (FK), which integrates joint angles down the kinematic chain.
However, the accuracy using FK depends on highly accurate motors and accumulates noise along the kinematic chain otherwise. 
Consequently, any pipelines requiring accurate 3D control and state estimation are typically demonstrated on highly expensive robot arms (usually over \$10K) \cite{shen2023F3RM,simeonovdu2021ndf,wilcox2025adapt3r}, and are difficult to deploy on lower-cost robots, which often have larger backlash and less precise encoder readings.

Other components of the 3D robotics stack are similarly restrictive and cumbersome. 
Camera-robot pose estimation requires an iterative approach which is tedious and importantly depends on highly accurate kinematic state estimation, similarly making it difficult for lower-cost hardware.
Robot calibration is likewise clunky, either depending on operators to match an approximate a ``target pose'' iteratively move to the minimum and maximum joint angles, or requires complex hardware with homing sequences \cite{spong2004robot,edition2005introduction}.
These requirements render the 3D robotics control stack slow, brittle, and inaccurate for lower-cost motors.

\begin{figure*}[thpb!]
  \centering
  \includegraphics[width=\textwidth]{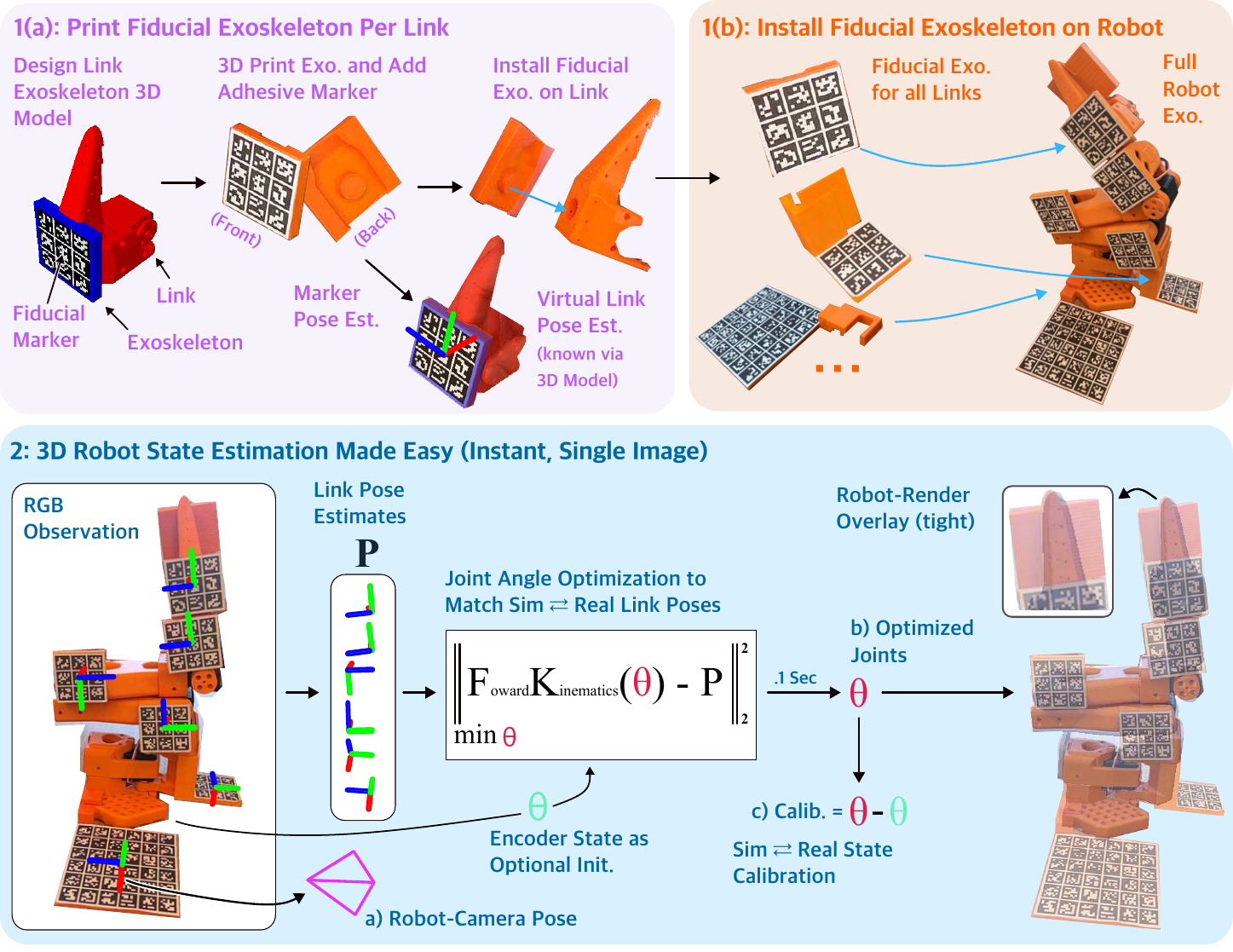}
   \caption{
    \textbf{Overview of using Fiducial Exoskeletons for robot state estimation.}  
    (1a, top-left) Each link is fitted with a \emph{fiducial exoskeleton}, a 3D-printed mount with a flat marker plane and a known marker-to-link coordinate transformation, enabling single-image 6D pose estimation for each link without any iterative calibration (Section \ref{text:fiducial_exo}).
    (1b, top-right) The exoskeleton for each link is printed and mounted on the robot.
    (2, bottom) From an RGB image, we estimate all link poses, also immediately yielding the camera pose directly from the base's marker [2a].  
    A fast optimization recovers the joint angles [2b] which best match the observed link poses (Section \ref{text:state_recovery}). 
    We also recover the robot's calibration offset [2c] by comparing the optimized joints to the raw encoder joints. 
    The final joint estimate tightly matches the physical robot, observed by the rendered robot overlay (rightmost).
    }
    \label{fig:overview}
\end{figure*}

To address these hardware and state-estimation challenges, we opt to use \emph{vision} for robot state estimation, which has exciting potential as it observes a ``holistic viewpoint'' over the full robot state, instead of just integrating per-motor observations.
In this work, we make vision a first-class signal for robot state estimation and control.
We reformulate the entire 3D estimation and control stack around 6D pose estimation of each link from an RGB image and a simple optimization to recover joint angles.
From the same set of poses, we also recover the robot to camera pose directly as well as robot calibration.
We further make control more precise by using state-based refinement: after naively moving to the target state, we estimate the robot state from vision, compute the delta to the target state, and finally move the robot with the delta direction.

To facilitate per-link pose estimation, we introduce the \emph{fiducial exoskeleton}: a lightweight 3D-printed mount with a fiducial marker \cite{garrido2014aruco,olson2011apriltag} for each link, providing an unambiguous marker-to-link transformation. The fiducial exoskeleton enables us to perform simple 6D pose estimation of each link from a single RGB image and without iterative calibration procedures which depend on the internal kinematics of the robot.

In summary, we introduce the following contributions:
\begin{itemize}
\setlength\itemsep{0em}
    \item A vision-centric reformulation of robot state estimation and control based on 6D link pose estimates, enabling recovery of joint angles, camera extrinsics, and robot calibration, from a single image and with significantly increased precision. 
    \item A practical mechanism --  the fiducial exoskeleton -- which simplifies per-link pose estimation.
    \item On a low-cost 6-DoF arm, our method reduces end-effector state estimation and control error by $\sim$75\% and  $\sim$45\%, respectively, over traditional forward-kinematics based estimation.
\end{itemize}
On a \$100 robot arm (SO-100 \cite{cadene2024lerobot}), our approach enables highly accurate state estimation compared to that from forward kinematics, as well as high-precision control.
\begin{figure}[thpb]
  \centering
  \includegraphics[width=.48\textwidth]{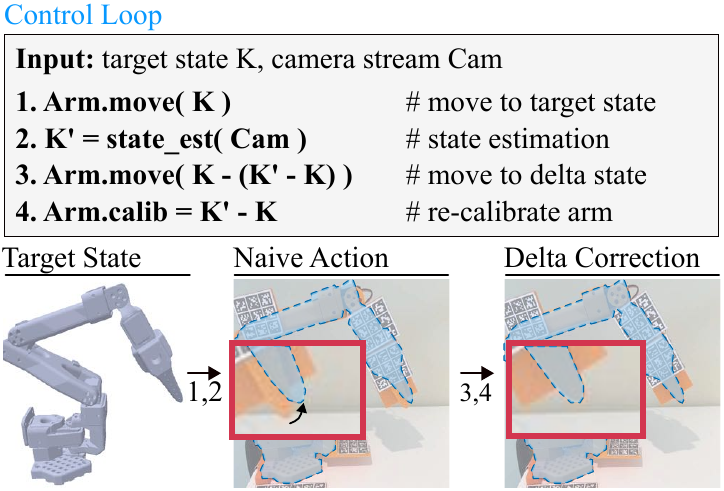}
  \caption{
    \textbf{The pseudo-code for our visual state-estimating control}.  
    Top: the pseudo-code of our control loop, where we first move the arm naively to the target position, estimate the current robot state, move the delta between the observed robot and target robot state, and finally re-calibrate the robot offsets for the next target motion.  
    Bottom: Illustrated steps of the control loop with the target state (Left), then the naive motion execution (Middle), and lastly the state-based refinement to better match the target state.  
    We also highlight insets on the end-effector to better emphasize the differences between the target and physical states.
    }
    \label{fig:control_loop}
\end{figure}

\begin{figure*}[thpb!]
  \centering
  \includegraphics[width=\textwidth]{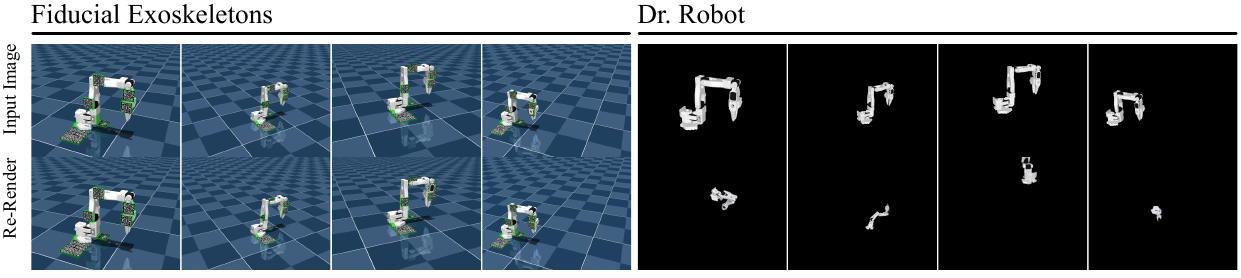}

\caption{\textbf{Comparison to Dr.~Robot on robot-camera pose and joint state estimation.}
We compare our (left) robot-camera pose and joint state estimates to those from Dr. Robot \cite{liu2024differentiablerobotrendering} (right), a rendering-based robot state estimation method, visualizing the input image (top row) and the re-rendered image (bottom row) using the inferred joints and camera pose. Dr. Robot's rendering is visualized using their internal splatting renderer and ours using the simulator renderer with our inferred parameters. Dr.~Robot infers pose and joints parameters  using differentiable rendering, which is known to be sensitive to local minima and initialization; its recovered pose is frequently misaligned with the input whereas our inferred parameters align near pixel-perfect with the input.
}
\label{fig:fidex_vs_drrobot}
  
\end{figure*}

\begin{figure*}[thpb!]
  \centering
  \includegraphics[width=\textwidth]{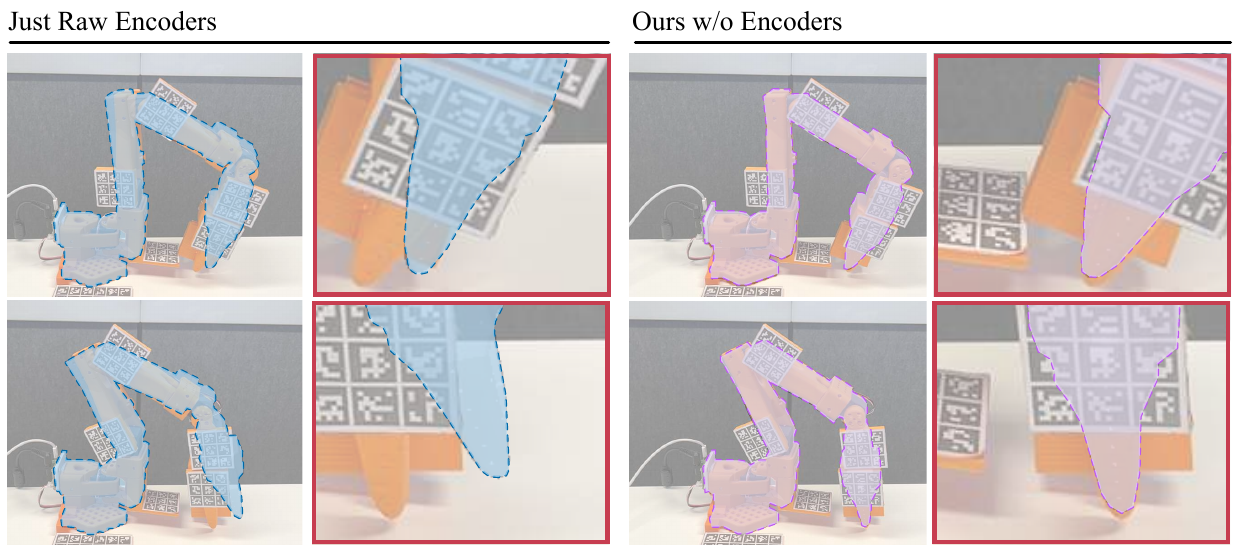}

    \caption{
       \textbf{State estimation results.}
       (Left) Using raw encoder readings for state estimation (left), the rendered-robot is not well aligned with the physical robot. 
        Using the fiducial exoskeletons, even without using any encoder readings as input, the robot re-render is aligned with the physical robot (Right).
        For each method, we plot the full-image overlay (inner left) and a highlighted inset on the end-effector (inner right).
    }
    \label{fig:state_results}

\end{figure*}

\section{Related Work}

\subsection{Classical Camera Pose Estimation}

The robot-camera pose is an important component in many robotic pipelines, linking the image space to the robot 3D workspace.
The standard way to obtain the camera pose is through hand-eye calibration \cite{Horaud_1995}, an iterative procedure where a robot operator attaches a fiducial marker on the end-effector and moves the robot through several kinematic states while capturing image observations.
The pose is recovered via an $AX=XB$ optimization~\cite{lenz1989calibrating,Horaud_1995}.
Note that this iterative data collection and the $AX=XB$ optimization is required only because the relationship between the fiducial marker and the end-effector is unknown. 
This hand-eye calibration is iterative, considered tedious to set up, has to be performed each time the robot or camera moves, and critically depends on accurate joint kinematics in order to provide correct motion pairs to the solver.  
Also note that this pipeline is not only tedious but difficult and inaccessible for low-cost robots with potentially noisy encoders.

In contrast, we simplify this design greatly with our fiducial exoskeleton by utilizing 3D-printed mounts where the relationship between the marker and the link is known, circumventing iterative data collection and the $AX=XB$ optimization.

\subsection{Robot Keypoint Prediction for Camera Pose Estimation}
Recent learning-based approaches aim to recover the camera pose from a single image, typically by predicting 2D keypoints of link centers \cite{lee2020cameratorobotposeestimationsingle,goswami2025robopeppvisionbasedrobotpose,tian2024robokeygenrobotposejoint,lu2023markerlesscameratorobotposeestimation,lu2024ctrnetxcameratorobotposeestimation,labbé2021singleviewrobotposejoint,ban2024realtimeholisticrobotpose}.  
These approaches then solve a PnP optimization from the 2D keypoints to the 3D model obtained from the forward kinematics and known joint states.
While reducing operator effort in avoiding the fiducial marker mounting and collection of multiple image observations, these methods similarly require highly accurate joint information, are only demonstrated on high-end robots ($>\$10$k) with precise encoders, have questionable generalization, and have awkward processing of occluded links.
In contrast, because our formulation predicts a full 6D pose for each link rather than sparse 2D locations, our method is able to not only recover the joint angles as well as the extrinsic calibration of external cameras with respect to the robot.

\subsection{Robot Pose Estimation from Vision}
Earlier work also explored markerless robot arm pose estimation with similar motivations of circumventing potentially unreliable kinematic state.
These approaches aimed to recover joint angles but often involved complex pipelines, often leveraging depth maps and heuristic methods for link clustering and segmentation \cite{bilić2023distancegeometricmethodrecoveringrobot,widmaier2016robot,zuo2025cravescontrollingroboticarm}. 
They also importantly do not recover the robot to camera pose.
In contrast, our approach simplifies these designs greatly without using any training data, while offering not just joint recovery but also camera pose estimation.

More recently, differentiable rendering approaches such as \emph{Dr. Robot}~\cite{liu2024differentiablerobotrendering} attempt to estimate robot state by optimizing the parameters of a differentiable robot appearance model (e.g., Gaussian splatting~\cite{kerbl20233d,nerf_survey_2024,remondino_nerf_critique_2023}) to match input images.  
These methods are powerful for enabling image-space-to-robot gradients and have been proposed as a way to integrate predictions from large vision models into robot pipelines.  
However, they require accurate robot segmentation, are sensitive to lighting and appearance conditions, and optimize in the raw RGB space -- making them slow, unconstrained, and often unstable.  
In contrast, our method leverages vision \emph{not at the pixel level} but at the structured level of per-link 6D poses, yielding a much more constrained, low-dimensional, real-time, and accurate optimization for robot state estimation.  

\subsection{Latent Inverse Robot Dynamics from Videos}
An orthogonal line of work aims to learn robot actions directly from video, often inferring \emph{latent} actions to explain the observed visual trajectories \cite{li2025unifiedvideoactionmodel,dreamgen_nvidia,collins2025amplifyactionlessmotionpriors,jang2025dreamgen,ye2025latentactionpretrainingvideos,playslot_icml2025,tian2024predictiveinversedynamicsmodels,wen2024vidmanexploitingimplicitdynamics}.  
For instance, DreamGen proposes a multi-stage pipeline that fine-tunes a video generator with tele-operated video data to generate novel robot data, without leveraging explicit robot actions, and an inverse dynamics model or latent action model to label trajectories.
Our method, in contrast, aims to recover instantaneous and explicit robot state.
These video-based latent dynamics models are thus complementary to our method: they provide implicit priors for action generation and policies, whereas our method offers explicit and reliable state estimation, perhaps to be used in tandem.

\section{Accurate State Estimation and Control of Robots using Fiducial Exoskeletons}
Our aim is to estimate the full 3D state of the robot (including the robot base to camera pose as well as the joint parameters) from a single RGB image, without depending on the robot's internal motor readings (which we consider here to be potentially noisy and unreliable). 
To achieve this goal of estimating the robot state from vision alone, we propose to first estimate the 6D pose of each link and then perform a global optimization over joint states to match the observed link poses. We also introduce the \emph{fiducial exoskeleton} to facilitate link pose estimation, which is effectively a 3D-printed attachment for each link with a fiducial marker. 
Below, we first describe the global optimization to recover joint states from link pose estimates \ref{text:state_recovery}, then how we estimate the poses of links using fiducial exoskeletons \ref{text:fiducial_exo}, and conclude with how we leverage this state estimation to control the robot with precision \ref{text:control}.

\begin{table*}[t]
    \centering
    
    \begin{minipage}{0.5\linewidth}
    \centering
    \begin{tabular}{lccc}
    \toprule
    & \multicolumn{3}{c}{State Estimation} \\
    \cmidrule(lr){2-4}
    Method        & Mask-IoU $\uparrow$ & Eff. Trans $\downarrow$ & Eff. Rot $\downarrow$ \\
    \midrule
    Ours Enc.     & \textbf{.85} & \textbf{.06} & \textbf{.27} \\
    Ours No-Enc.  & .84 & .07 & .35 \\
    Just Enc.     & .78 & .18 & 1.1 \\
    \bottomrule
    \end{tabular}
    \end{minipage}%
    \hfill
    \begin{minipage}{0.5\linewidth}
    \centering
    \begin{tabular}{lccc}
    \toprule
    & \multicolumn{3}{c}{Control} \\
    \cmidrule(lr){2-4}
    Method        & Mask-IoU $\uparrow$ & Eff. Trans $\downarrow$ & Eff. Rot $\downarrow$ \\
    \midrule
    Ours w/ Delta   & \textbf{.79} & \textbf{.21} & \textbf{2.6} \\
    Ours w/o Delta & .78 & .22 & 2.8 \\
    Naive Move     & .63 & .37 & 3.8 \\
    \bottomrule
    \end{tabular}
    \end{minipage}
    \caption{State estimation (Left) and control (Right) results with fiducial exoskeletons are far stronger than using encoder-readings alone. For both state estimation and control, we compare the re-rendered robot using the estimated sensor state to the ground-truth robot mask (Mask IoU), as well as the rotation and translation distance for the estimated and true end-effector position. For state estimation, we compare the raw-encoder state (Just Enc.) to our method without using encoder states as initialization (Ours No-Enc.) and then with using the encoder states as well (Ours Enc.). For control, we compare the target robot state to the naive encoder motion (Naive Move), ours without the delta refinement (Ours w/o Delta), and with the delta refinement (Ours w/ Delta).}
    \label{tab:state_quant}
\end{table*}

\subsection{Recovery of Robot Joint States from Link Poses}
\label{text:state_recovery}

Visual estimation of the 6D pose for each link provides direct constraints on the joint parameters of the robot: by comparing the link poses induced by forward kinematics to those observed from vision, we can optimize the joint parameters to match the visual observations.
Forward kinematics induces a set of per-link poses 
$ \{ T_j(\theta) \}_{j=1}^L$ by integrating the link transformations and joint angles down the kinematic chain. That is, for a $d$-DOF robot with joint parameters $\theta = (\theta_1, \ldots, \theta_d)$, 
\begin{equation}
T_j(\theta) = \prod_{i=1}^{j} \,T_i^{i-1}(\theta_i) ,
\end{equation}
where $\,T_i^{i-1}(\theta_i) \in SE(3)$ is the transformation of joint $i$ with respect to its parent frame $i-1$. Note that the forward kinematics integration is differentiable with respect to the joint states, an important property which we will leverage below.

In parallel, we assume access to a visual estimate of the same set of link poses,  
\begin{equation}
\mathcal{P}_{\text{obs}} = \{ P^{\text{obs}}_j \}_{j=1}^L ,
\end{equation}
where each $P^{\text{obs}}_j \in SE(3)$ is the 6D pose of link $j$.

We can then \textit{solve} for the optimal joint states which best aligns the observed link poses to the ones induced by forward kinematics:
\begin{equation}
\theta^\star = \arg\min_{\theta} \sum_{j=1}^L d\!\left( T_j(\theta), \, P^{\text{obs}}_j \right)^2 ,
\end{equation}
where $d(\cdot,\cdot)$ is a distance metric on $SE(3)$.
While this optimization is nonlinear, it is low-dimensional and can be solved with off-the-shelf nonlinear solvers (e.g. L-BFGS \cite{liu1989lbfgs}) in real-time.
Also, note that we can leverage encoder readings when available as the initial $\theta$ value used in the optimization, and just initialize the joint values with zeros otherwise.
We find that the joint-state optimization typically converges to the same values with or without the encoder readings as initialization, and only observe an improvement in using the encoder readings as initialization when most links are occluded.

Also note that while the robot base is not technically a link, we assume the robot base pose is also estimated in this set of observed poses, and transform all link poses to the robot coordinate frame. In the next section, we also describe estimating the robot base pose with respect to external cameras.

\captionsetup[table]{labelfont=normalfont,textfont=normalfont}

\subsection{Link Pose Estimation via Fiducial Exoskeletons}
\label{text:fiducial_exo}

Above we illustrated how we can leverage 6D link poses to estimate the state of the robot. 
While there are indeed several potential parameterizations of the per-link pose regression, here we highlight another direction of \emph{simply attaching a fiducial marker to each robot link}: fiducial markers provide SE(3) marker-to-camera pose estimates from a single RGB image, robust to partial occlusions and generalize trivially to a wide distribution of orientations.
The key difficulty here, however, lies in registering the marker coordinate frame to the coordinate frame of the link they represent: while the standard approach of registering markers to the robot frame is the classical and multi-observation $AX=XB$ calibration, this assumes access to accurate joint kinematics and a known camera pose. 
In the case of imprecise motors, accurate kinematic state or camera pose is difficult to obtain for such calibration. 

To remedy these issues, we introduce the \emph{fiducial exoskeleton}. 
For each link, we 3D-model a lightweight 3D-printed mount that attaches unambiguously onto the link. 
The fiducial exoskeleton consists of two parts: the component which mounts to the link, and a flat plane which holds the fiducial marker (see Fig. \ref{fig:overview} for illustration). 
Importantly, this 3D model provides us the direct transformation between the plane to the link coordinate frame, which allows us to recover the 6D pose of the link from the 6D marker estimate, without any iterative calibration, kinematic state, or camera pose.
Formally, the link pose in the camera frame is simply:  
\begin{equation}
T^{\text{cam}}_{\text{link}} \;=\; 
T^{\text{cam}}_{\text{aruco}} \;\; T^{\text{aruco}}_{\text{exo}} \;\; T^{\text{exo}}_{\text{link}},
\end{equation}where $T^{\text{cam}}_{\text{aruco}}$ is the pose of the fiducial marker in the camera frame,  
$T^{\text{aruco}}_{\text{exo}}$ is the transform from the marker to the exoskeleton,  

\begin{figure*}[thpb]
  \centering
  \includegraphics[width=\textwidth]{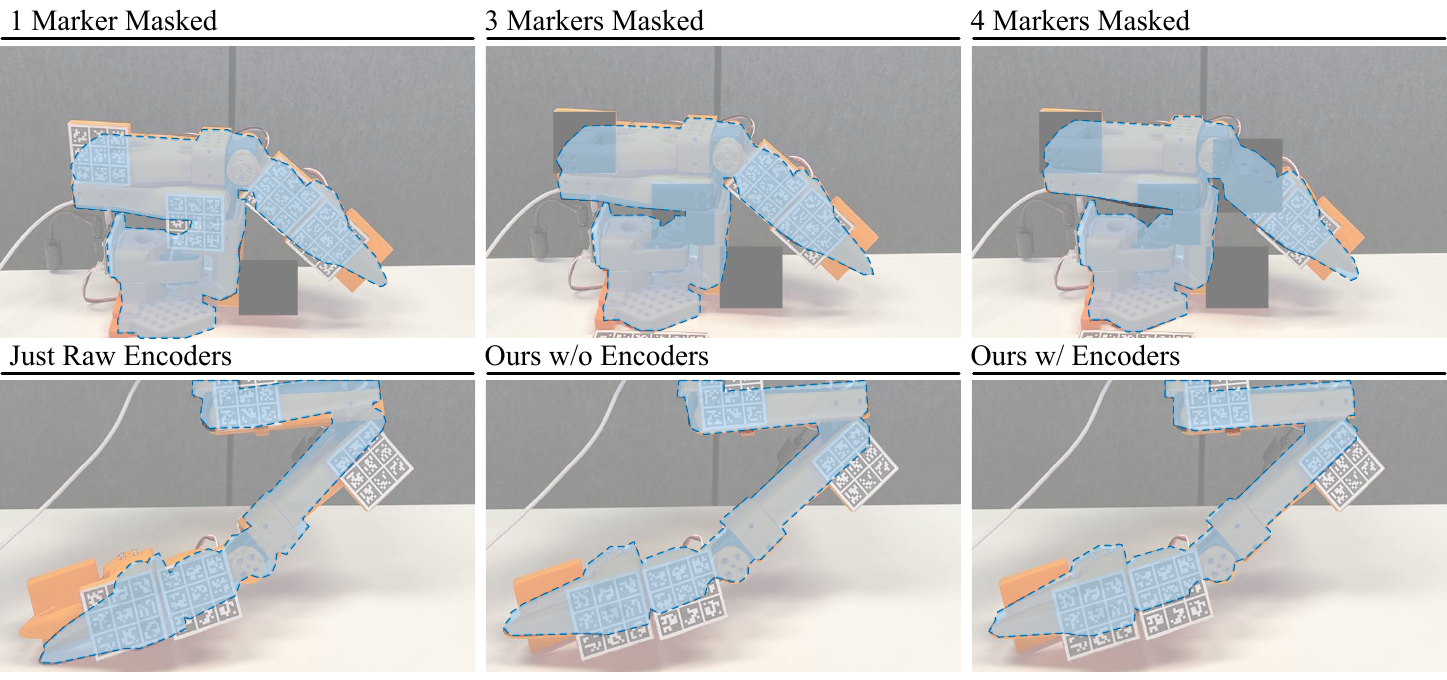}
  \caption{
    \textbf{Robustness study}.  
    (Top) State estimation results when masking out $1$, $3$, and $4$ fiducial markers.
    Even with all but one marker occluded, our vision-based state estimation still produces strong results. 
    Even with significant occlusion, our optimization remains stable, with encoder initialization helping in the most under-constrained cases.  
    (Bottom) Unlike learning-based approaches which may struggle with out-of-distribution robot configurations, here we show state estimation even in an `upside-down' state is similarly accurate for our method's estimation.
    }
    \label{fig:robustness}
\end{figure*}

\begin{table}[t]
\centering

\begin{tabular}{lcc}
\toprule
\textbf{Metric} & \textbf{FidEx (Ours)} & \textbf{Dr.~Robot} \\
\midrule
Joints (L2)                 & \textbf{0.0855} & 3.3852 \\
Camera Translation (m)      & \textbf{0.0053} & 0.7392 \\
Camera Rotation ($^\circ$)  & \textbf{0.2148} & 109.36 \\
Time (s)                    & \textbf{0.0427} & 4.9311 \\
\end{tabular}
\caption{Quantitative comparison of state estimation (joints and robot-camera pose) accuracy with Dr. Robot \cite{liu2024differentiablerobotrendering}. 
Lower is better for all metrics. ``Time'' denotes optimization time to perform inference.}
\label{tab:fidexvsdrrobot}
\end{table}

\begin{table}[!]
    \centering
    \begin{tabular}{lcccc}
    \toprule
    Method        & 1 Occ. & 3 Occ. & 4 Occ. & Upside Down \\
    \midrule
    Ours (Enc)    & \textbf{.88} & \textbf{.87} & \textbf{.87} & \textbf{.82}  \\
    Ours (No-Enc) & \textbf{.88} & \textbf{.87} & .75 & \textbf{.82}  \\
    Just Enc      & .81 & .81 & .81 & .76  \\
    \bottomrule
    \end{tabular}
    \caption{State estimation results (re-rendered robot mask IoU) under varying marker occlusions and robot orientation conditions. Note how even without any sensor information (Ours No-Enc), our optimization only falls into a degenerate minima when only the end-effector is visible.
    }
    \label{tab:marker_occ}
    \end{table}

\subsection{Camera Robot Pose Estimation and Robot Calibration from a Single Image for Free} 

\subsubsection{Camera Robot Pose}

In the same way we attach fiducial exoskeletons to each link, we attach one to the robot base as well. 
This immediately delivers robot-to-camera extrinsics, rendering it as simple as estimating the pose of the base's fiducial marker:
\begin{equation}
T^{\text{robot}}_{\text{cam}} \;=\; 
(T^{\text{cam}}_{\text{aruco}} \;\; T^{\text{aruco}}_{\text{exo}} \;\; T^{\text{exo}}_{\text{robot}})^{-1}.
\end{equation}
This direct estimation eliminates the need for tedious and iterative hand-eye calibration procedures.

\subsubsection{Robot Calibration}

Also note that since our joint estimation does not require known encoder states, we can perform robot calibration from a single image as well. The joint offset is recovered as the difference of the raw encoder joints $\theta^{Enc}$ and the optimized joints $\theta^\star$:
\begin{equation}
\Delta\theta = \theta^\star - \theta^{Enc} .
\end{equation}

Note this calibration procedure is performed from a single image and is dramatically simpler than current calibration procedures, which often involve manually moving the robot to a `target pose' for alignment or require iterative and slow homing procedures.

\begin{figure*}[h!]
  \centering
  \includegraphics[width=\textwidth]{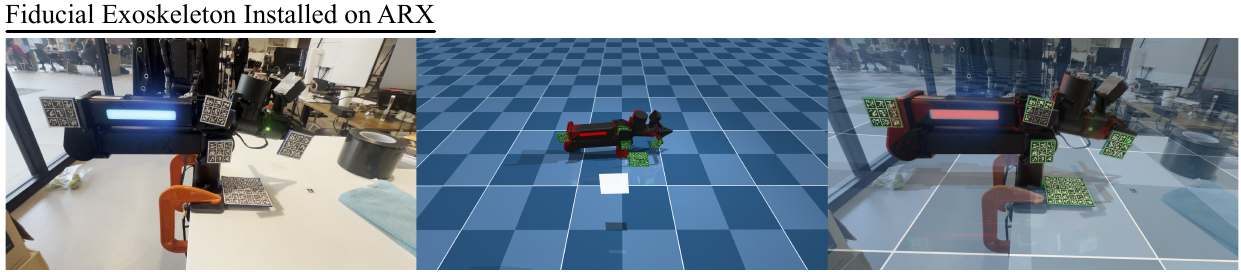}
   \caption{
    \textbf{FidEx on other embodiments. Fiducial exoskeletons can be installed on a wide range of robot embodiments; here we show installation and single-image state recovery on the ARX arm, with additional platforms to be supported in future work. }  
    }
    \label{fig:challenges}
\end{figure*}

\begin{figure}[h!]
  \centering
  \includegraphics[width=.5\textwidth]{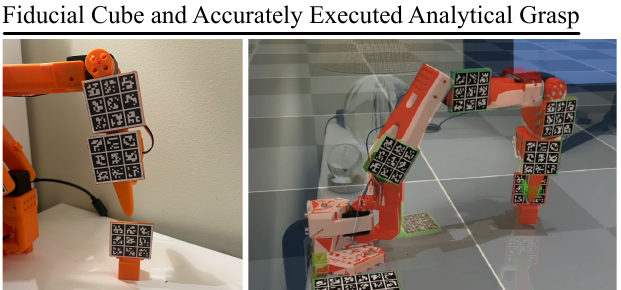}
   \caption{
    \textbf{Fiducial cube and analytical grasp with FidEx. Left: our 3D-printed ArUco fiducial cube and robot setup, used to evaluate object grasp success with an analytical IK grasp as a proxy task for calibration accuracy. We compare grasp success under our FidEx single-image calibration versus a common approximate manual (“middle-pose”) calibration. Right: an example analytical grasp execution, showing the planned pose overlay in white and the rendered robot in red, illustrating the tight alignment achieved with our calibration.}  
    }
    \label{fig:challenges}
\end{figure}

\subsection{Accurate Control via State Estimation Refinement}
\label{text:control}

For the same reason that state estimation of a robot with potentially inaccurate sensors can be difficult -- local regions of backlash, drift, etc. -- commanding the robot to a target kinematic state  $\theta^{Targ}$ with precision can be difficult as well.  

We introduce a simple control loop to leverage our state estimation for pose refinement. 
For each kinematic target, we perform on-the-fly robot calibration, command the robot to the target state $\theta^{Targ}$, estimate the current robot state $\theta^{*'}$, and move the robot with the delta between the observed and target state $\theta^{Targ}-(\theta^{*'}-\theta^{Targ})$. 

The procedure pseudocode and visual results are described in Fig.~\ref{fig:control_loop}. 
With this control loop, we are able to greatly increase the precision at which we can move the robot to a target kinematic state compared to naive execution without requiring high-precision encoders.

\section{Experiments}

\begin{figure*}[thpb!]
  \centering
  \includegraphics[width=\textwidth]{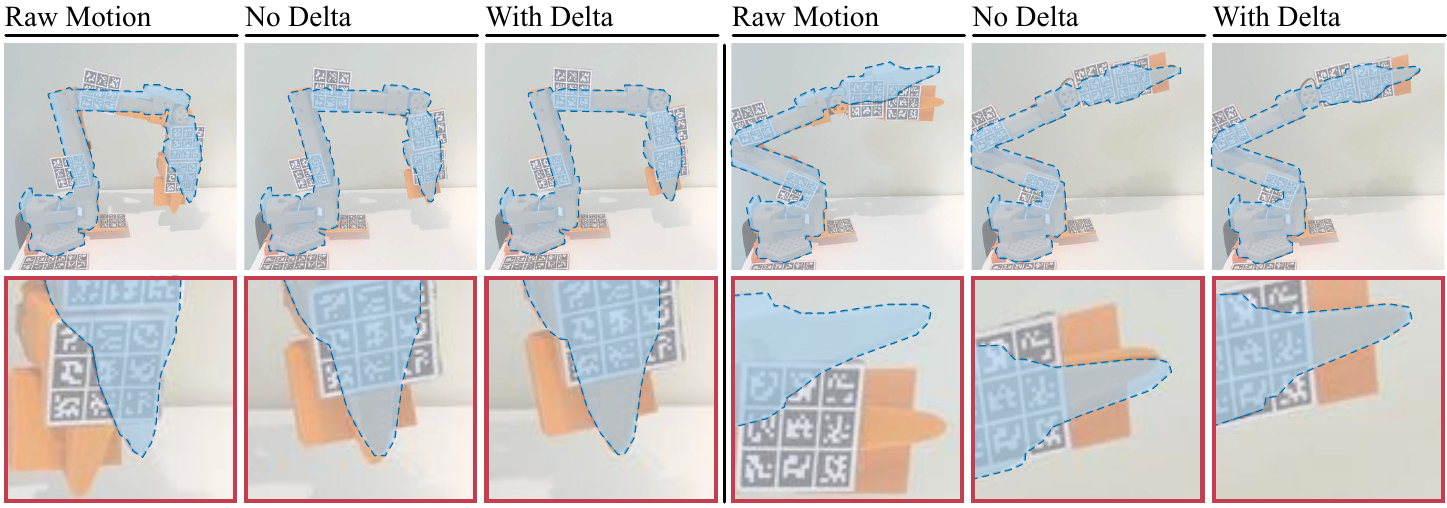}
  \caption{
    \textbf{Robot control illustrations}.
    For each target state (blue outline) we compare the difference to the physical robot in using just the raw naive motion encoders (Raw Motion), the execution with our fiducial exoskeleton but without our delta refinement (No Delta), and finally with our delta refinement (With Delta). 
    We also plot highlighted insets on the end-effector (Bottom). 
    Without delta correction, the end-effector location is still close to the target, and the delta refinement further closes the gap between the target and executed motion.
    }
    \label{fig:control_results}
\end{figure*}

We benchmark our method in both robot state estimation and control on a low-cost robot in Figures \ref{fig:state_results},\ref{fig:control_results} and Tables \ref{tab:state_quant},\ref{tab:marker_occ}. 
For state estimation, we evaluate how well we can estimate the joint parameters and 6D link positions in \ref{exp:state}, and for control \ref{exp:control}, how accurately we can move the robot links to target kinematic states or 6D link positions.

\subsection{Estimating Robot State}
\label{exp:state}
For estimating the robot state, we compare the traditional approach of using FK (`Just Enc.') for the 6D position of each link, to our method -- both with using the raw encoder states as input (Ours Enc.) and then without (Ours No-Enc).
To compare the state estimation, we plot the robot mesh render overlayed on the robot in Fig. \ref{fig:state_results}, and report Mask IoU as well as the 6D pose error (position and rotation losses) for the end-effector, on a dataset of diverse robot configurations, in Table \ref{tab:state_quant}.
Our method produces state estimation which are much more aligned with the physical robot than that of naively using FK-based integration. 
See Fig. \ref{fig:state_results} to qualitatively observe how much more aligned the robot re-rendering is with the physical robot.
And quantitatively, see Tab. \ref{tab:state_quant} where we report a $\sim$75\% decrease in error in estimating the end-effector's SE(3) position.

We additionally compare against Dr.~Robot \cite{liu2024differentiablerobotrendering}, a recent method that estimates robot state via differentiable rendering, optimizing robot-camera pose and joint parameters to match an observed image.
In contrast to FidEx, which leverages fiducial markers (making the visual background largely irrelevant), Dr.~Robot relies on a splatting-based rendering optimization, which makes the optimization sensitive to initialization and local minima.
We plot qualitative results in Fig.~\ref{fig:fidex_vs_drrobot}, showing the input image and the corresponding re-render using the recovered camera and joint parameters.
While FidEx produces near pixel-perfect alignment between the input and re-render, Dr.~Robot often converges to substantially misaligned solutions, reflecting the difficulty of solving full pose and joints recovery purely via rendering-based optimization.
Quantitatively, Table~\ref{tab:fidexvsdrrobot} shows that FidEx achieves dramatically lower joint and camera errors while running over two orders of magnitude faster (e.g., $97.5\%$ lower joint error, $99.3\%$ lower camera translation error, $99.8\%$ lower camera rotation error, and $99.1\%$ lower runtime).

\subsection{Robustness Studies}
One reasonable question is how our method performs when some subset of the links are occluded, or the robot is in `out-of-distribution' positions.
We cover various numbers of link markers (1, 3 and 4 markers), as well as move the robot to an upside-down position, and report accuracy in Table \ref{tab:marker_occ} and plot illustrations in Figure \ref{fig:robustness}.
When all links are occluded, our method reduces to the case of just using the raw encoder states, and when even one link (such as the end-effector) is visible, even this single constraint increases the state estimation accuracy significantly.
We only find that in the case of not using any encoder state (not relevant in practice and more so a demonstration of surprising robustness) and only one link is visible, our estimation can fall into a degenerate minima.
And since our state estimation does not involve any dataset or learning, there is no notion of out-of-distribution pose estimation: our pose estimation reduces to that of fiducial marker estimation, a classical method with robust implementation, and so this `upside-down' state is trivially recovered as well. 
See Fig. \ref{fig:robustness} (bottom) to observe that our estimation in this `upside-down' configuration is similarly aligned with the physical robot.

\begin{figure}[thpb]
  \centering
  \includegraphics[width=.5\textwidth]{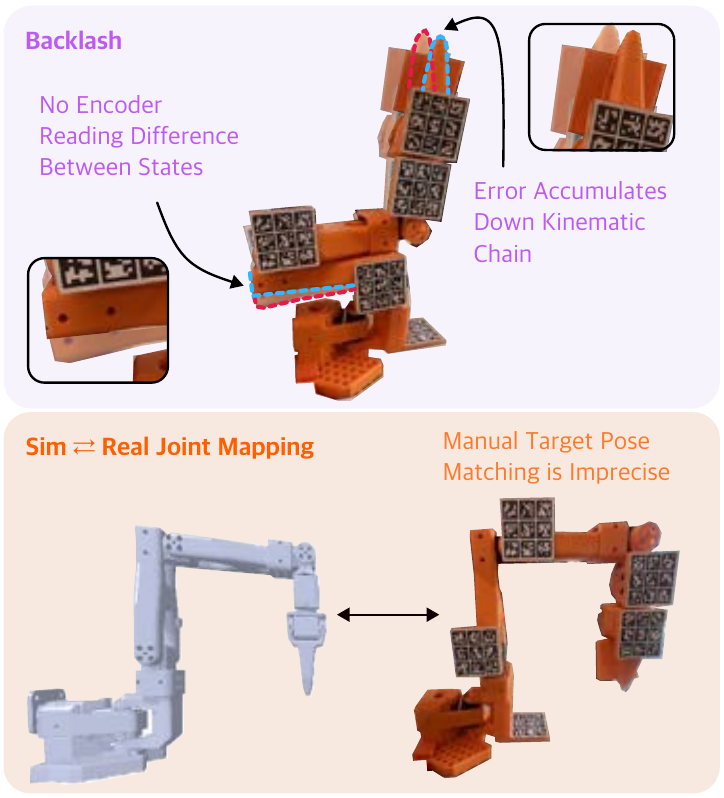}
   \caption{
    \textbf{Challenges of 3D state estimation on low-cost robots.}  
    (\textbf{Top}) \emph{Backlash:} due to imperfect mechanical alignment in the internal motors, each joint has a margin of rotation in which encoders register no change.  
    This unmeasured change accumulates down the kinematic chain, leading to large state error at the end-effector.
    (\textbf{Bottom}) \emph{Calibration:} joint offsets are often calibrated by manually matching one or a small set of `target' poses, yielding imperfect mappings.
    }
    \label{fig:challenges}
\end{figure}

\subsection{Control Precision Studies}
\label{exp:control}
To measure how precisely we can move the robot to a target 6D and kinematic state, we record a diverse dataset of target positions, and move the robot to each target state using our control loop.
We similarly plot the robot mesh re-render of the target state onto the final reached state in Fig. \ref{fig:control_results}, as well as quantitative Mask IoU and end-effector 6D position accuracy, in Table \ref{tab:state_quant}.
We find that each component of the control loop is vital for accurate positioning, comparing our full method to the naive method of just moving the robot to the target state, as well as not using our state-estimating delta position refinement.
Specifically, we find that even without using the delta-correction (effectively just re-calibrating the robot before moving), our method reduces the end-effector control error by $\sim$40\%, and adding our delta-correction further decreases error by an additional $\sim$5\%.

We also evaluate grasping precision on a small 3D-printed puck with an ArUco marker to indicate the target grasp location, using inverse kinematics to plan an analytical grasp, first executed with our FidEx single-image calibration and the common middle-pose alignment. 
With FidEx, the robot successfully grasped the puck in 9 out of 10 trials, whereas the middle-pose calibration only succeeded in 4 out of 10 trials, highlighting the improved control accuracy enabled by our method.

\section{Discussion}

We have introduced Fiducial Exoskeletons, a simple and robust design for robot state estimation and control which does not depend on highly accurate and expensive motors, instead leveraging holistic image-based estimation.
Our method is significantly more accurate the standard method (forward kinematics with raw sensor encoders) for robot state estimation and control on a low-cost 6DoF robot arm.
We believe our design paves the way towards simpler and more accurate 3D robot state estimation and control, as well as increases capabilities for lower-cost robot arms.

\subsection{Limitations}

Fiducial Exoskeletons have several limitations that suggest exciting future explorations. 
First, the robot has to be observed by an external camera in which at least the base marker and one other marker are clearly visible: exploring the camera mount design to ensure proper visibility is an interesting constraint and future design consideration. 
One such solution could be to use one mounted camera just focused on observing the robot state and another for the robot and the rest of the scene together.
The markers also have to be designed per-embodiment and per-link, which is laborious but is a constant design time per embodiment and can be distributed to all robot operators via 3D CAD files. 
And while aesthetics are not the primary concern of future robot policy learning, the markers are large and occlude much of the robot; future work on integrating more seamless and subtle marker design is interesting as well.

While there are indeed physical design limitations here, we hope this direction of 3D robot co-design increases the breadth of robots by which we can leverage analytical robot control and 3D-based policy learning, especially in the lower-cost regime of robots with less precise motors.

{
    \small
    \bibliographystyle{ieeetr}
    \bibliography{main}
}

\end{document}